\documentclass[letterpaper, 10 pt, conference]{ieeeconf} 
\IEEEoverridecommandlockouts 
\makeatletter
\def\endthebibliography{%
	\def\@noitemerr{\@latex@warning{Empty `thebibliography' environment}}%
	\endlist
}
\makeatother
\usepackage{graphicx}
\usepackage{amssymb}
\usepackage{amsmath}
\usepackage{setspace}
\usepackage{ulem}
\usepackage{mathrsfs}
\usepackage{url}

\usepackage[dvipsnames,table]{xcolor} 

\usepackage{algorithmic}
\usepackage{algorithm}
\usepackage{tikz}
\newcommand*\circled[1]{\tikz[baseline=(char.base)]{
		\node[shape=circle,draw,inner sep=2pt] (char) {#1};}}

\graphicspath{{./fig/}}

\title{\LARGE \bf A Low-cost Robot with Autonomous Recharge and Navigation for Weed Control in Fields with Narrow Row Spacing}

\author{Yayun Du$^{1}$, Bhrugu Mallajosyula$^{1}$, Deming Sun$^{1}$, Jingyi Chen$^{1}$, Zihang Zhao$^{1}$ \\ 
Mukhlesur Rahman$^{2}$,Mohiuddin Quadir$^{3}$,
Mohammad Khalid Jawed$^{1}$ 
\thanks{$^{1}$Department of Mechanical \& Aerospace Engineering, University of California, Los Angeles, 420 Westwood Plaza, Los Angeles, CA 90095}%
\thanks{$^{2}$Department of Plant Sciences, North Dakota State University, Fargo, 1340 Administration Ave, ND 58108.}%
\thanks{$^{3}$Department of Coatings and Polymeric Materials, North Dakota State University, Fargo, 1340 Administration Ave, ND 58108.}
}

\begin{document}

\maketitle

\begin{abstract}
	Modern herbicide application in agricultural settings typically relies on either large scale sprayers that dispense herbicide over crops and weeds alike or portable sprayers that require labor intensive manual operation. The former method results in overuse of herbicide and reduction in crop yield while the latter is often untenable in large scale operations. This paper presents the first fully autonomous robot for weed management for row crops capable of computer vision based navigation, weed detection, complete field coverage, and automatic recharge for under \$400. The target application is autonomous inter-row weed control in crop fields, e.g. flax and canola, where the spacing between croplines is as small as one foot. The proposed robot is small enough to pass between croplines at all stages of plant growth while detecting weeds and spraying herbicide. A recharging system incorporates newly designed robotic hardware, a ramp, a robotic charging arm, and a mobile charging station. An integrated vision algorithm is employed to assist with charger alignment effectively. Combined, they enable the robot to work continuously in the field without access to electricity. In addition, a color-based contour algorithm combined with preprocessing techniques is applied for robust navigation relying on the input from the onboard monocular camera. Incorporating such compact robots into farms could help automate weed control, even during late stages of growth, and reduce herbicide use by targeting weeds with precision. The robotic platform is field-tested in the flaxseed fields of North Dakota.
\end{abstract}


\section{Introduction}
\label{sec:introduction}
Weed management is one of the most important tasks in any agricultural operation. Left unmanaged, weeds compete with crops for water, soil nutrients, and sunlight, hindering plant growth and reducing crop yield~\cite{weedinfluence}. As such, technological innovation in the weed spraying market is in the economic interest of farmers and the general public alike. Nonetheless, due to heavy use of herbicides, the environment is polluted~\cite{lewis2009herbicides} and weed resistance is increasingly aggravated~\cite{weedresistance}. Therefore, methods of applying herbicide precisely are gaining traction within the robotics community.

Precision farming and localized weed control are methods of reducing herbicide use while effectively managing weed populations in farms. There has been an extensive amount of research in robotics for precision agriculture~\cite{slaughter2008autonomous, perez2015highlights, chebrolu2019robot, guo2018multi}. However, robust technologies are lacking for autonomous weed management in row crops with narrow line spacing, e.g. flaxseed and canola where the spacing between two croplines can be only one foot. Adding to the challenge, access to reliable electricity is typically unavailable in agricultural fields. North Dakota, where we tested our robot, contributes 91\% of U.S. flax production; U.S. market for flaxseeds oils in 2020 is $\$0.12$ billion~\cite{flaxMarketSize}. Effective and low-cost methods for weed control in narrow crop rows can be particularly beneficial to the economy of the state of North Dakota.

\begin{figure}[h!]
	\centering
	\includegraphics[width=0.8\columnwidth]{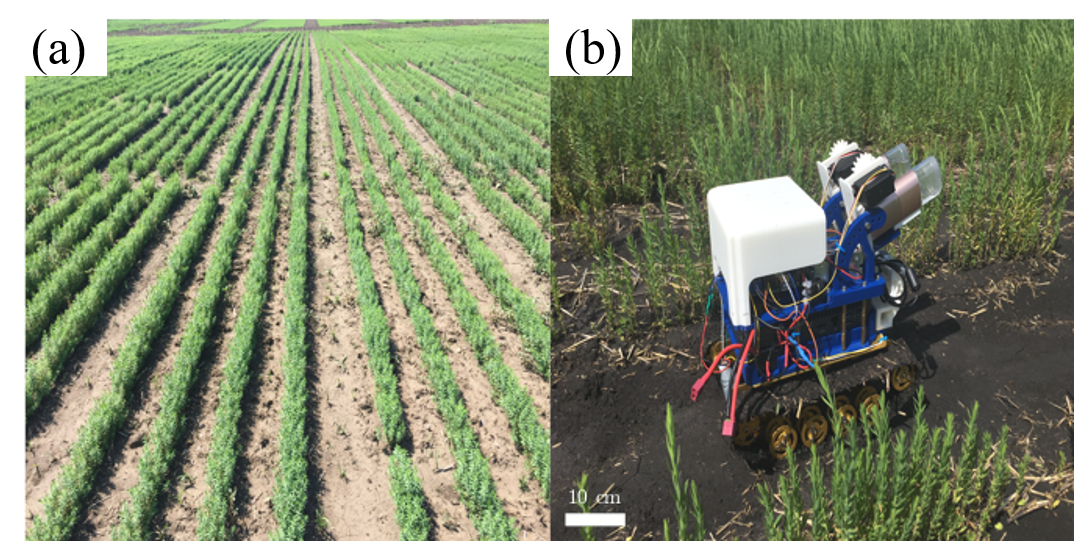}
	\caption{(a) Flaxseed fields in Fargo, North Dakota, where the robot was tested. (b) Photograph of the agriculture robot working in the field.
	}
	\label{fig:overview}
\end{figure}

Though precision farming is highly effective, methods of implementing precision farming show room for improvement. Existing guided sprayers are large tractors~\cite{perez2015highlights, gonzalez2017fleets} and even recent commercialized robots, e.g. Ecorobotix that sprays small doses of herbicide to kill weeds, are still meter-sized. They are generally equipped with high-precision but rather expensive real-time kinematics (RTK) GPS~\cite{chebrolu2019robot} or Multi-Global Navigation Satellite System (GNSS) for navigation to move along pre-planned routes~\cite{guo2018multi}. In addition, they are suitable only for large farms and cannot spray weeds with sufficient accuracy after crop canopies have been established. As a result, functional compact robots able to work within small corridors between croplines are needed. In case of a compact cost-effective robot, the aforementioned navigation methods are neither affordable with acceptable accuracy nor feasible due to restrictions on size. Instead, a monocular camera is a much more economical and functional option for both navigation and weed identification. Monocular camera based computer vision weed spraying robots have two main tasks: crop row guidance and differentiation of weeds from crops. Guerrero \textit{et al.} summarized and described the available crop row guidance algorithms~\cite{guerrero2013automatic}:(a) exploration of horizontal strips, (b) stereo-based approach, (c) Hough transformation, (d) vanishing point-base, (e) blob analysis, (f) accumulation of green plant, (g) frequency analysis and (h) linear regression. Nonetheless, if weed density is high, line fitting methods, e.g. Hough transform, could be slow because of the large number of line extraction and error-prone due to incorrectly fitted lines. Other methods either rely on more expensive sensors than cameras or are of high computational complexity and are therefore not applicable. Power endurance is another concern for autonomous robots due to limited space for batteries on the robot, calling for the exploration of outdoor recharging methods. However, current recharging methods only apply to indoor scenarios~\cite{cassinis2005docking, zebrowski2005recharging}. Even the best commercially available robotic lawn mowers only take care of lawns and return to indoor recharging station with the help of GPS or boundary wire.

This paper presents the design of an autonomous computer vision based weed spraying robot small enough to travel between croplines during all stages of growth. It is equipped with multiple herbicides and capable of inter-row navigation via fusing information from Inertial Measurement Unit (IMU) and camera (see Figs.~\ref{fig:overview} and ~\ref{fig:hardware}). Due to the need to work in narrow crop spacing, our robot is small with a unique camera perspective. This is the first one, to our knowledge, without an overhead view of fields, limiting its navigation options as mentioned above. Next, based on the established hardware framework, several inter-row navigation algorithms depending on monocular camera vision are explored. Line detection algorithms and feature point extraction method are applied but they turn out to be computationally expensive and unrobust. As a result, we developed a navigation method, \textit{Pre-Contour-Gradient}, a color-based contour algorithm combined with special preprocessing (Pre) techniques, enabling the robot to navigate not only between straight but also curved and irregular croplines. The robot makes decision to turn to the next line by detecting the decrease in area (Gradient) while reaching the end of each cropline. 
Moreover, the robot is capable of returning to the charging station and charging itself when low battery is detected. In summary, our main contributions are as follows. 

\begin{itemize}
    \item We designed and built the first low-cost but experimentally-validated robotic framework capable of performing fully autonomous weed control in flaxseed and canola fields. The angle-adjustable spraying and recharging systems are two innovations in hardware design.
    \item A vision-based algorithm, \textit{Pre-Contour-Gradient}, is used in conjunction with IMU enabling fast and robust inter-row navigation and turning for complete full field exploration
    \item Software and hardware solutions for autonomous recharging are presented. A computer vision algorithm is presented to help align the charging arm with the socket, allowing the robot to get charged safely and operate continuously in the fields.
\end{itemize}
The paper is organized as follows. 
We explain our hardware innovations in Section \ref{sec:hardware} mainly consisting of a newly designed spraying system and a robust recharging system, followed by Section \ref{normalpath} in which our novel inter-row navigation algorithm is illustrated. Then, the experimental setup and test results in real flaxseed fields in North Dakota and testbed on campus are displayed in Section \ref{sec:experiment}. In the last section, concluding remarks and future focus are drawn. \par

\section{Hardware Innovation}
\label{sec:hardware}
\begin{figure}[t!]
	\centering
	\includegraphics[width=\columnwidth]{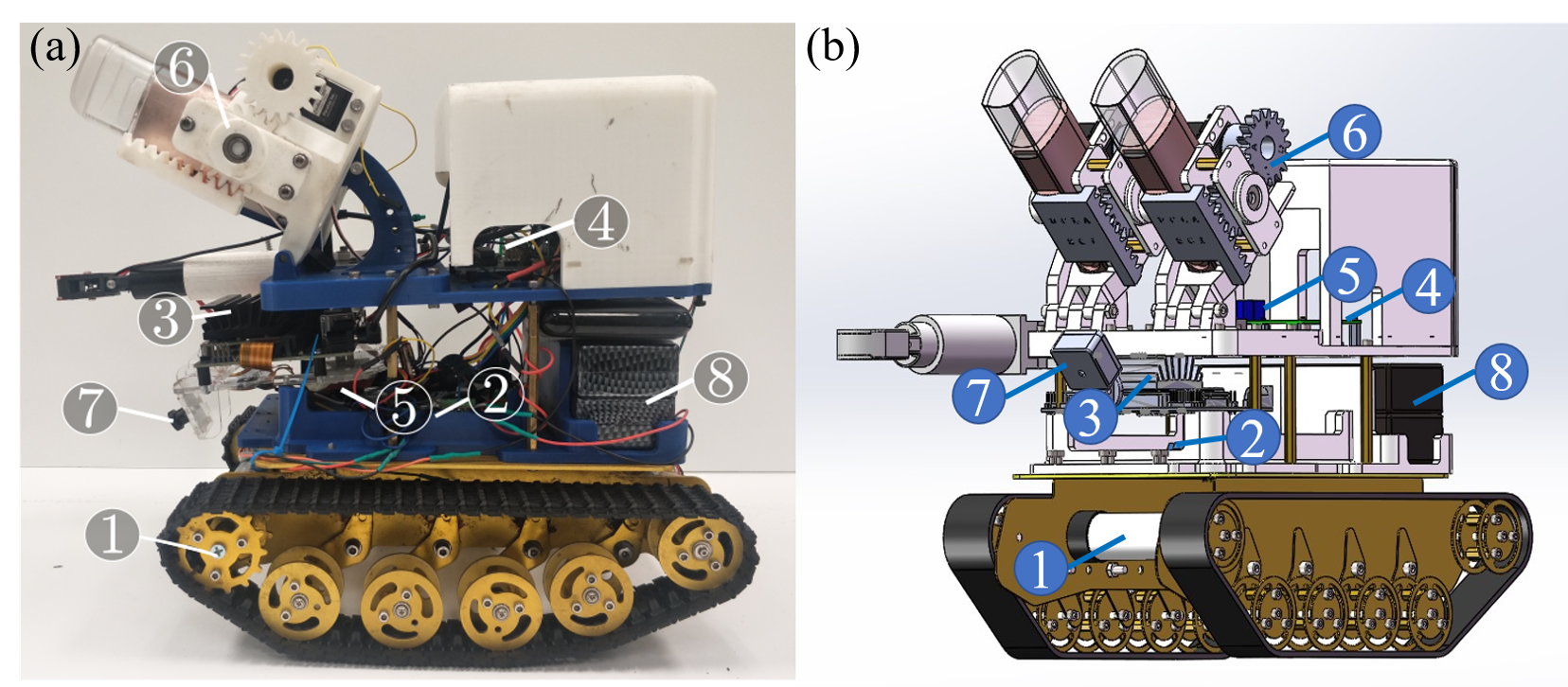}
	\caption{Hardware layout of the agriculture robot. (a) Actual robot used for experiments. (b) The robot design in SOLIDWORKS computer-aided design software.
	}
	\label{fig:hardware} 
\end{figure}

Referring to Fig.~\ref{fig:hardware}, the robot fits within a small form factor, just 36 cm long, 21 cm wide, and 37 cm tall with a spring suspension. The drivetrain consists of two \circled{1} DC motors (labeled as \circled{1} in Fig.~\ref{fig:hardware}) rated at 3.0 kg-cm torque with encoders integrated.
The motors are connected to an \circled{2}  Adafruit Motorshield v2 motor controller which is, in turn, connected to the \circled{3} NVIDIA Jetson Nano that serves as the main controller for the robot, low cost but with efficient image processing.
The  encoders  of  the motor are connected to an \circled{4} Arduino Uno. A \circled{5} BNO055 Absolute Orientation Sensor is connected to Jetson Nano.  This  IMU measures  and computes the orientation of the robot during navigation. Attached to the very front of the chassis is a \circled{7} 60 FPS, 136 field-of-view (FOV) wide-angle CSI camera with an unobstructed  view of  the  path  before  the  robot. As shown in Fig. \ref{fig:hardware}, this camera is connected to the Jetson Nano with an adjustable perspective, $0^\circ \leq \theta \leq 120^\circ$ relative to the vertical line, 12cm above the ground. 
Apart from electronics that need to be easily disassembled and strictly ventilated, others are enclosed in a 3-D printed plastic shell, which protects the electronics from contamination by the dusty environment, entanglement with the stretching crop canopies, and erosion due to the high air humidity in rainy weather.

\begin{figure}[ht!]
	\centering
	\includegraphics[width=0.8\columnwidth]{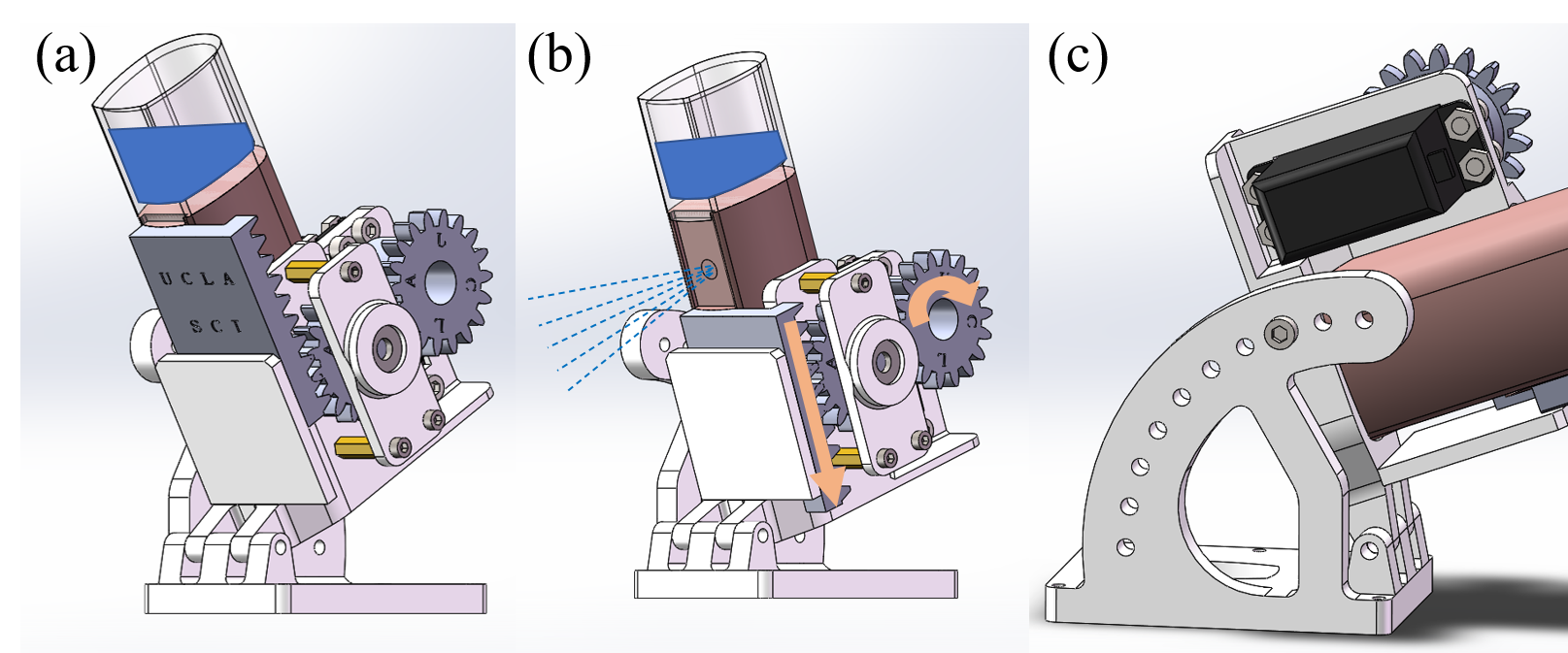}
	\caption{Spraying system when (a) turned off and (b) turned on. (c) Close-up of sprayer holder from side view of (a).
	}
	\label{fig:spraying}
\end{figure}

\subsection{Spraying System}
Because of the narrow distance between row crops, off-the-shelf sprayers with long hose controlled by solenoid valves do not function on our robot for precise spraying. Mounted near the front of the robot are the two \circled{6} herbicide spraying assemblies. 
As designed, the relative position of sprayers to the camera is considered accounting for two main aspects. As seen in Fig.\ref{fig:hardware}, the herbicide sprayed from the sprayer nozzle will not be spilled onto the camera. The angle of the sprayers in Fig. \ref{fig:spraying} is also adjustable by screwing the bolt into different holes drilled on the 3-D printed part as shown in Fig. \ref{fig:spraying}(c). The number of spraying assemblies can be extended for multi-weed classification in the future. Each spraying assembly, shown in Fig.~\ref{fig:spraying}, consists of an electronic mister filled with a 25ml of liquid herbicide, a rack and pinion-based system to activate and deactivate the misters, and a servo with position feedback. The servo is connected to and controlled by the Arduino Uno. 

\begin{figure}[hb!]
	\centering
	\includegraphics[width=0.7\columnwidth]{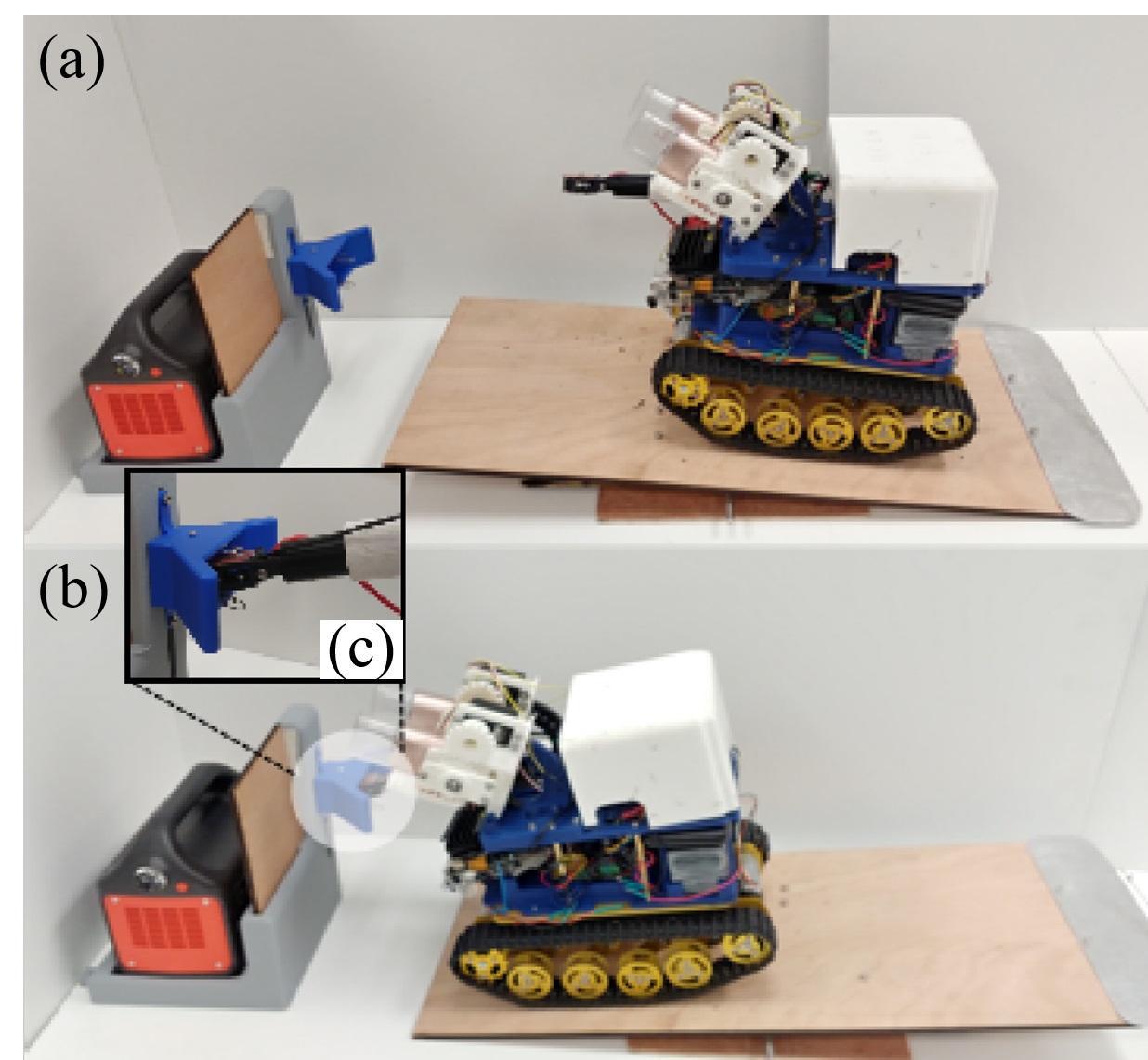}
	\caption{(a) The robot climbs to the ramp to recharge. (b) The robot passes the midpoint of the ramp and the charging arm contact the conductive terminal on the charging station. (c) Close-up of the charger design. Note: the springs in the charging arm are compressed slightly for a seamless contact.}
	\label{fig:recharge}
\end{figure}

\subsection{Vision \& Recharging System}
\label{subsec:vision}

Our recharging mechanism relies on a portable energy-saving hardware design and a corresponding vision-based charging station detection algorithm.

\subsubsection{Recharging hardware system}
\label{subsubsec:rechargingSys}

Extending from the front of the robot is one 3-D printed arm with conductive metal pads at its end. The metal pads are connected to the charging terminals of \circled{8} the two 8.4V lithium polymer batteries on the robot. Referring to Fig.~\ref{fig:recharge}, when the metal pads make contact with the conductive terminals of the charging station, the batteries begin to charge. 

A wooden ramp is placed before the charging station such that when the robot passes the midpoint of the ramp, it tilts and gravity pulls the robot towards the charging station, solidifying electrical contact while consuming no extra electricity. A solar generator can be extended for autonomous operation of the robot over several weeks. On the 3-D printed case, there is a three red dot pattern attached, which in conjunction with the funnel-shaped charging header that functions as a homing head and the computer vision algorithm allows the robot to align the robot charging arm with the charging header.
\subsubsection{Vision-based Charging Station Detection}
\label{subsubsec:rechargVision}
When batteries get low, the robot first navigates back along the original path until it is around 3 meters away from the charging station. Next, onboard vision is used again for perfect alignment between the charging arm and socket during recharging as presented in Fig. \ref{fig:recharge}. 
A three evenly-spaced red dot pattern (in Fig. \ref{fig:exp_circleDetection}) with the center of the center circle being the target point is designed to guide the robot. 
This color and pattern are selected as the HSV value range of red is very stable during regular working hours under various weather conditions, 139-172 (H), 22-143 (S), 154-255 (V) facing the sun and 0-255 (H), 0-89 (S), 61-255 (V) under backlighting.
In addition, the environmental disturbance of circle detection is not as much as rectangles or other straight-arm polygons.  If the charging arm is 2-4cm off, the funnel-shaped header will still lead it to successful charging.


Once the centers of circles are recognized, the camera should be aligned with the target point, the centroid of circle centers. The algorithm used to detect the circle is named as \textit{Def-circle} by us. The pseudo code of \textit{Def-circle} is given in Algorithm 1. \textit{Def-circle} algorithm is essentially a combination of utilization of mathematical definition of a circle, color-based filtering, contour detection, and circle Hough Transform.
The main steps are briefly illustrated below.\\
(1) Contrast Limited Adaptive Histogram Equalization (CLAHE) is adopted to remove blur caused by the glare and increase color contrast first;\\
(2) According to the mathematical definition of a circle, all points in the same plane lying at an equal distance from a center point are determined to lie on a circle contour;\\
(3) Wrongly detected circles that are far from the blue charging funnel-shaped header should be removed;\\
(4) If the approach of mathematical definition of circles does not find potential candidates,  circle Hough Transform embedded in OpenCV is applied to detect circles within the neighborhood of the blue charging head.

\begin{algorithm}[ht]
\caption{Pseudo code of \textit{Def-circle} algorithm}
\begin{algorithmic}[1]
	\FOR{\textbf{I} in video streaming} 
	\STATE $\texttt{I}$ $\gets$ Apply \textit{CLAHE} to each channel in RGB space\;
    \STATE $\texttt{I}$ $\gets$ \textit{Resized} and \textit{binarized}  by \textit{threshold} 160-200\;
    \STATE $\texttt{C}$  $\gets$ Find \textit{ vertices}  of contours in $\texttt{I}$ \;
    \FOR{i from 1 to $len$($\texttt{C}$)}
		\IF {$len(\texttt{C}_i) > minPts$}
			\STATE  Find the center, $\texttt{cn}$, and radius, $\texttt{rd}$, of minimum enclosing circle of $\texttt{C}_i$ \;
			
			\STATE $\texttt{ofs}$ $\gets$ Compute the variance of distance between points on $\texttt{C}_i$ and $\texttt{cn}$ \;
			
			\IF {$\texttt{ofs} < \texttt{maxOfs}$ $\And$ $\texttt{rd} >\texttt{minR}$ $\And$ $\texttt{rd} \leq \texttt{maxR}$ }  
			\STATE $\texttt{p}$ $\gets$ [int($\texttt{cn}$), int($\texttt{rd}$)] \;
			\STATE Append $\texttt{p}$ to $\texttt{Res}$
			\ENDIF
		\ENDIF
 			\IF{$len(\texttt{Res}) > 0 $}
			\FOR{j from 1 to $len$($\texttt{Res}$)}
			\IF{Area of red neighbors of $\texttt{Res}_j$ $>$ $\texttt{thres}$}
			\STATE Append $\texttt{Res}_j$ to $\texttt{Res\_N}$ 
			\ENDIF
			\ENDFOR
			\ELSE
			\STATE	$\texttt{Res\_N}$ $\gets$ Use \textit{HoughCircles} to find the circles \;
			\IF{$len(\texttt{Res}) > 0 $}
			\STATE Perform the same steps in lines 15-19
		\ELSE
		\STATE Move forward straight
		\ENDIF
		\ENDIF
	\ENDFOR
	\ENDFOR
\label{algo::defCircle}
 \end{algorithmic}
\end{algorithm}


Our method has typical parameters, such as minimum number of points on the contour to be decided as a circle further ($numPts$), maximum offset from the point on the contour to the contour center ($\texttt{maxOfs}$), minimum and maximum radius of the minimum enclosing circle ($minR$ and $maxR$), and the threshold helpful to remove the false positives ($thres$). These parameters can be made adaptive later. 
When the batteries are fully charged, the robot backs up and withdraws from the ramp with the aid of gravity after passing the midpoint of the ramp. 
The accuracy of circle detection of our method is 20\% higher than the circle Hough Transform method which holds an accuracy of about 50\%. Detailed experimental results in fields can be found later in Section \ref{subsec:rechargingExp}.

Overall, the effective and robust recharging system is an ideal coordination of hardware and software. The former includes a wooden ramp that utilizes gravity to decrease energy dissipation, a mobile station, and a circle pattern aiding the alignment of charging arm and socket. The latter involves a circle pattern recognition algorithm, \textit{Def-circle}.

\section{Inter-row Navigation}
\label{normalpath}
\begin{figure}[h!]
	\centering
	\includegraphics[width=\columnwidth]{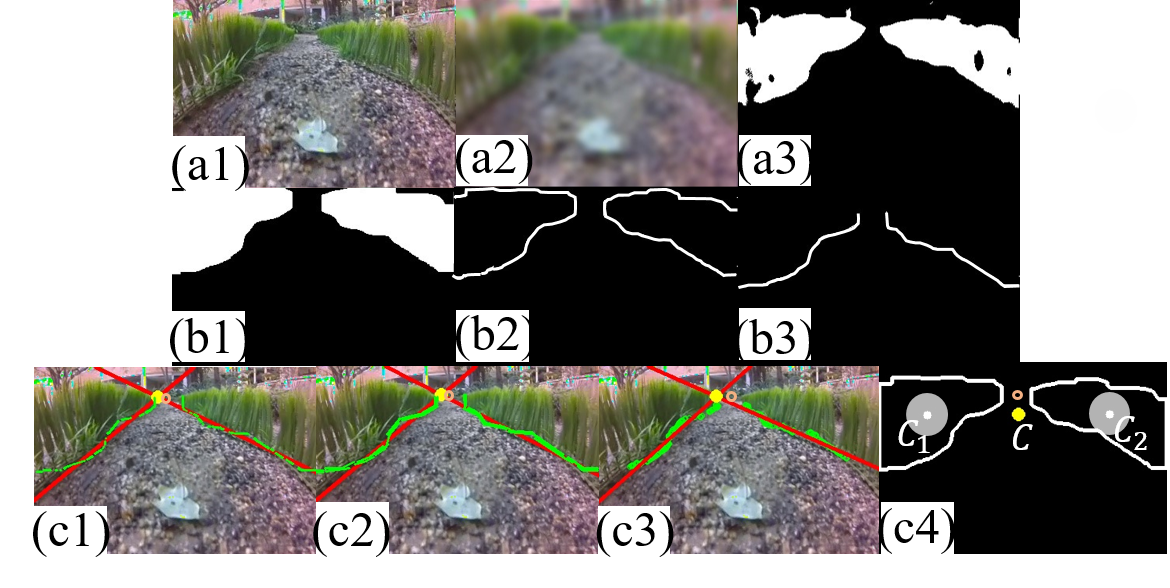}
	\caption{Overall image processing procedure with (a) Image preprocessing and (b) algorithms applied to find target point. (a1) Original image. (a2-a3)(b1-b3) Gaussian blur, HSV color filtering, erosion and dilation, canny edge detection, and region of interest cropping applied, in turn. (c1-c3) PHT, LSD, and FPE applied. (c4) Contour algorithm and centroids. Centroid of the left contour is denoted as $C_1$, centroid of the right contour is $C_2$, and the centroid of the whole contours is $C$. The hollow circle in (c1-c4) denotes the ground truth of the vanishing point that we researchers label.
	}
	\label{fig:preprocess}
\end{figure}

The first subsystem controls  the  movement  of the  robot  from  one  end of a row to  the  other, avoiding collision  with  the  croplines  that  circumscribe  its path. For the algorithms and experiments described later, the frame sequences were captured with resolution of 360×240 pixels. The image processing is divided into two phases: preprocessing and target localization. In Figs.~\ref{fig:preprocess}(a2-a3) and (b1), we combine noise reduction, edge detection, color-based segmentation, and morphology to preprocess the image. Gaussian Blur, HSV color filtering, and  erosion and dilation are noise reduction techniques applied firstly to remove debris between the croplines. Next, canny edge detection is implemented to detect the edges (Fig.~\ref{fig:preprocess}(b2)). Finally, region of interest (ROI) cropping helps crop out the top of the crops while leaving their lower stem untouched. Image preprocessing makes navigation more computationally efficient as shown in Fig.~\ref{fig:preprocess}(b3). As for target point localization, we compared the vanishing point generated from different line detection algorithms including Probabilistic Hough transform (PHT)~\cite{kiryati1991probabilistic}, line segment detector (LSD)~\cite{von2012lsd}, and feature point extraction (FPE)~\cite{rublee2011orb} and the results are shown in Figs.~\ref{fig:preprocess}(c1-c3), respectively. Although PHT is an optimization of standard Hough Transform and is proved to have the highest accuracy in the methods above, it keeps the inherent deficiency of Hough Transform, i.e. its high computational cost, which results in significant delays while running on Jetson Nano board. We recall how people walk between croplines by ignoring crop-specific details such as spacing and periodicity and only focusing on avoiding green plants. This inspired us to employ contour algorithm with some preprocessing steps which will be introduced in the next subsection. 

\subsection{Image Preprocessing}
\label{subsec:ImagePreprocess}
\subsubsection{Complete Procedure}
\label{subsubsection:procedure}
The image outcome after every step mentioned above is shown in Fig.~\ref{fig:preprocess}. Firstly, we apply Gaussian Blur with a 25×25 sliding window to reduce the noise,
rocks on the path and inconsequential details such as texture about the plants. 
Comparing Fig.~\ref{fig:preprocess}(a2) with Fig.~\ref{fig:preprocess}(a1), note that details and noise such as stones and leaves on the path are all removed after Gaussian Blur. The robot camera will take pictures of fields upon entering, which will be converted to the hue, saturation, value (HSV) colorspace, more robust to varying illumination conditions compared with RGB color space. 
In Fig. \ref{fig:preprocess}, the upper and lower thresholds of green crop HSV range are (21, 43, 46) and (77, 255, 172) respectively. The pixels that fall into this range determined during initial calibration will be kept and all others removed. However, HSV value range of plants in fields varies depending on the conditions of illumination, varying with the movement of the sun during the day. In order to make the color-based navigation more robust to light conditions, we come up with an algorithm enabling the robot to adjust the HSV range automatically in Section \ref{subsubsec:self-adjust}. \par

The crop region extracted by HSV filter still contains numerous imperfections due to gaps between crops and the unevenness of roots as shown in Fig.~\ref{fig:preprocess}(a3). As such, opening and closing morphological operators used to find specific shapes in an image are applied~\cite{le2020novel}. Specifically, the opening operation helps smooth the contour in an image and remove small objects; it comprises of erosion followed by the dilation operation. On the other hand, the closing operation tends to eliminate small holes and fill gaps in the contours. We use a 21×21 square structuring element to input the opening and closing morphological operators for filtering. As Fig.~\ref{fig:preprocess}(b1) shows, the crop region is smoothed while boundaries are sharpened.
Fig. \ref{fig:preprocess}(b2) shows that canny edge detection helps find the outline of two crop regions. It can be seen from Fig.~\ref{fig:preprocess}(b3) that other edges apart from root lines are ruled out according to the slope based on human perspective. In summary, preprocessing is helpful for image downsampling, making computation more effective.
\begin{figure}[h!]
	\centering
	\includegraphics[width=0.7\columnwidth]{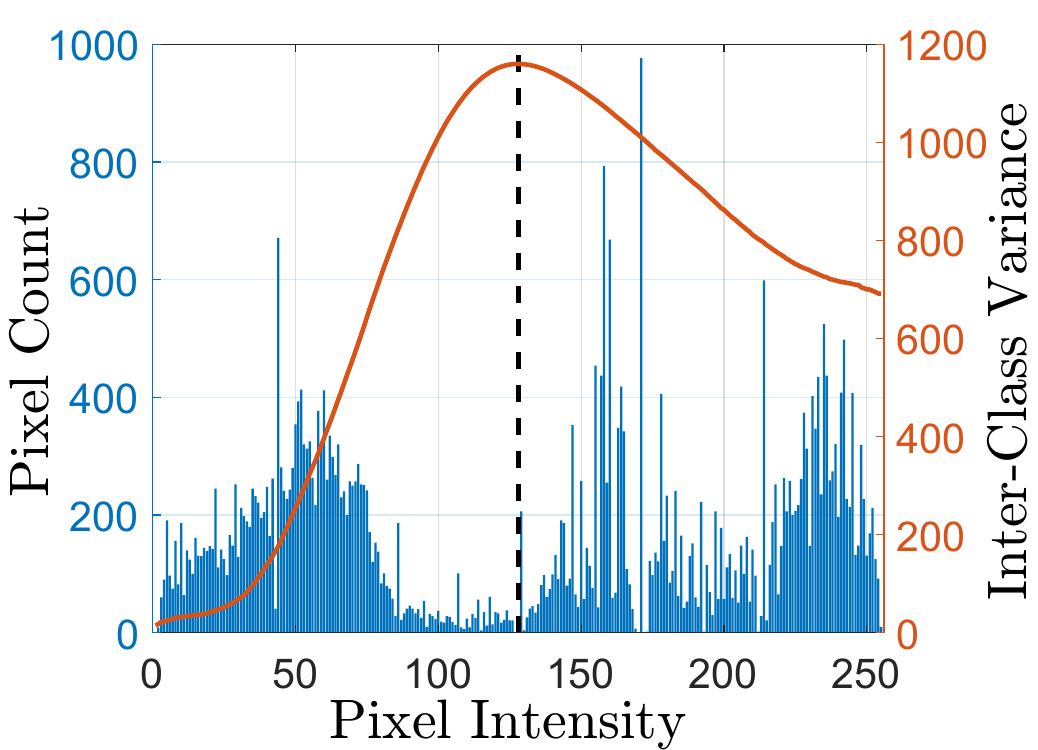}
	\caption{Figure showing how Otsu's method determines the threshold of H value to segment Fig. \ref{fig:preprocess}(a1). 
	}
	\label{fig:otsu}
\end{figure}

\subsubsection{Self-adjusting HSV Range}
\label{subsubsec:self-adjust}
Here, we name the new method that we used for the robot to self-adjust the HSV range as \textit{Prior-Otsu} method. Initial HSV range is set when the robot enters the fields before working, by dragging the trackbars inside graphical user interface (GUI) programmed with OpenCV computer vision programming library. This HSV range is then used by the contour algorithm in Section \ref{subsec:contour} for inter-row navigation during the first hour. The robot stores the coordinates of contour centroids (points $\mathbb{C}_1$ and $\mathbb{C}_2$ in Fig. \ref{fig:preprocess}(c4)) and the areas of corresponding contours, at ten different locations when its deviation angle is less than $5^\circ$ within the first hour of operation. As shown in Fig. \ref{fig:preprocess}(c4), circles with $\mathbb{C}_1$ and $\mathbb{C}_2$ being the center and with the area being $\frac{1}{5}$ of the average area of ten recorded contours are drawn. Fifty pixels are selected randomly within each of these two circles, the average of which the corresponding HSV values are calculated and saved. \par 
Specifically, hue (H) value dominates in deciding the accuracy of recognition of green plants and the change of it will not exceed 30 within an hour except extreme weather conditions according to our experiments from 7 a.m. to 6 p.m. in Summer in Fargo, North Dakota. However, considering the situation where the fifty pixels chosen are not representative, Otsu's method \cite{otsu1979threshold} is utilized when the change of H value is over the threshold, 30. 
As seen in Fig. \ref{fig:otsu}, Otsu's method divides the H value at every pixel into histograms, calculates the inter-class variance, and finds the desired threshold (the pixel intensity/H value where the black dashed line is located) to differentiate two classes, plants and soil. 
The self-adjusting HSV range module combines the \textit{prior} HSV range in the last hour with \textit{Otsu}'s method as a backup, and therefore named as \textit{Prior-Otsu} method. 


\subsection{Contour-Gradient Algorithm}
Two lines of crops, not necessarily straight, are filtered out after the overall preprocessing procedure. In this section, our target point localization algorithm -- contour algorithm -- is presented, the speed and accuracy of which are compared with three line detection algorithms (Probabilistic Hough Transform (PHT)~\cite{winterhalter2018crop}, LSD~\cite{von2012lsd}, and FPE incoporated with random sample consensus (RANSAC)) in Section \ref{sec:algorithm_result}.
\label{subsec:contour}  
Each pixel in the image is binarized after HSV color filtering introduced in Section \ref{subsec:ImagePreprocess}. Contour detection predicts the posterior probability of a boundary with an orientation at each image pixel $(x,y)$ by measuring the difference in local image brightness, color and  texture channels~\cite{martin2004learning}. Thus, boundary points, i.e. contours, between green plants and the environment will be extracted. With obstacles properly determined, the robot will have a good understanding of surrounding environment and walk through by following the centroid of the obstacles (crops). In this algorithm, image moment is calculated by 
\begin{equation}
M_{ij} = \sum_{x}\sum_{y}x_iy_jI(x,y),
\end{equation}
where $I(x,y)$ is the pixel intensity at coordinate $(x_i,y_j)$. Here, moments are applied to calculate the enclosed area and position of each contour. The area of binary images is the zeroth moment $M_{00}$ and thus centroid is evaluated as 
\begin{equation} \label{eq:contour}
\begin{split}
(x_c,y_c) &= (\frac{M_{10}}{M_{00}}, \frac{M_{01}}{M_{00}})       \\ &= (\frac{\sum_{x}\sum_{y}xI(x,y)}{\sum_{x}\sum_{y}I(x,y)}, \frac{\sum_{x}\sum_{y}yI(x,y)}{\sum_{x}\sum_{y}I(x,y)}).
\end{split}	
\end{equation}

In Figs. \ref{fig:preprocess}(c1-c3), the two straight lines are the two croplines extracted by applying corresponding three line detection algorithms, PHT, LSD, and FPE, from the root line segments obtained from Fig. \ref{fig:preprocess}(b3). The target point is derived as the intersection point of these two straight lines, which is shown as a filled circle in Figs. \ref{fig:preprocess}(c1-c3). In contrast, in Fig. \ref{fig:preprocess}(c4), the two white contours are generated by contour algorithm with $C_1$ and $C_2$ being the two centroids, the center of which is target point $C$. The yellow hollow circle in Figs. \ref{fig:preprocess}(c1-c4) stands for the ground truth of target point labelled by us researchers. 
%
Once the target point computation in the image coordinate system is complete using contour algorithm, the relative orientation of the target point to the current orientation of the robot measured by IMU can be determined. A PID control loop monitors the deviation between the desired and current orientation of the robot and returns the gains to the motor shield. In the end, the motors are controlled by the motor shield 
such that the deviation angle will always be controlled within $5^\circ$. 

Compared with the existing line extraction algorithm-oriented navigation methods, contour algorithm is proven to not only run faster with a higher accuracy between croplines.
Contour algorithm also guides the robot to turn to the next row when  the  end  of  a  row  is near, the area of greenness in the scope of the camera decreases rapidly, i.e. there is  a constant \textit{gradient} of the green area. If it drops under a threshold (2300 when the plants are 10cm with no big gaps between seedlings), the robot decides to turn $90^\circ$ first and then another $90^\circ$, and continue to march in the next line. 
 The overall algorithm for navigation is therefore called \textit{Pre-Contour-Gradient}. 

\begin{figure*}[t!]
	\centering
	\includegraphics[width=0.75\textwidth]{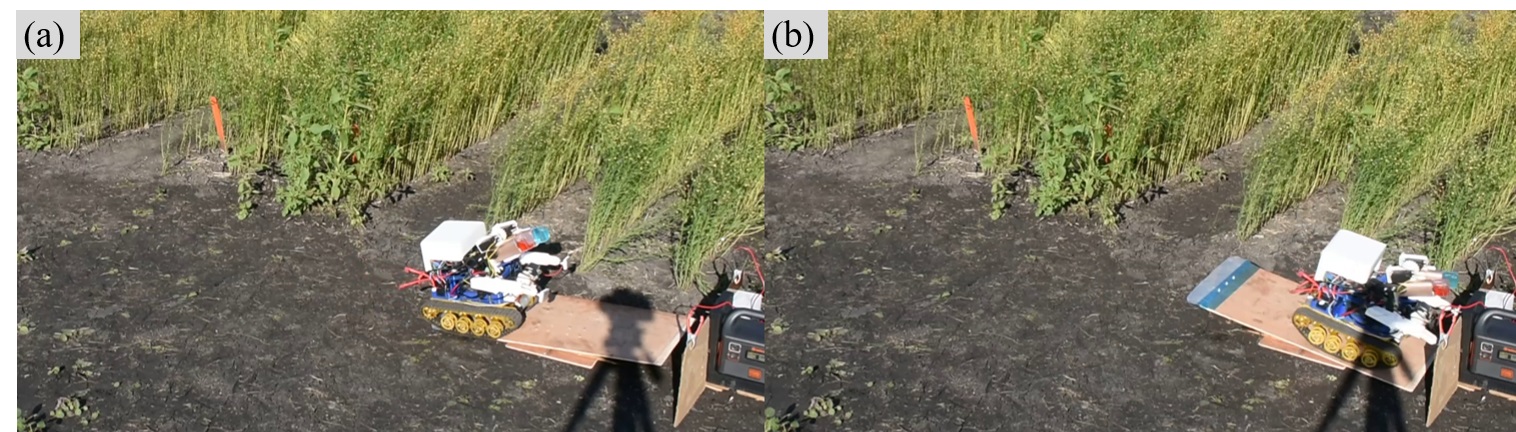}
	\caption{Images showing how the robot goes back to recharging station to recharge itself in the field. (a) The robot climbs to the ramp to recharge. (b) The robot passes the midpoint of the ramp and the charging arm contact the conductive terminal on the charging station.}
	\label{fig:ramp_in_fields}
\end{figure*}
\section{Experiments} 
\label{sec:experiment}
North Dakota is the leading producer of flax for oil and food use, with about 92\% of U.S. flax production and 88\% of U.S. acreage. The experiments were performed under different weather conditions -- sunny, cloudy, and windy weather over a one week period in August, 2019 on the flaxseed fields of North Dakota State University, Fargo, North Dakota. Example images from those fields are shown in Fig. \ref{fig:overview}. Our experiments were designed to verify that our robot is able to realize the functions mentioned in Section~\ref{sec:introduction} in real fields. 35 tests were conducted, each field test consisting of operation between 6 adjacent $3$m long rows. In between these 6 rows, the width of the aisles ranges from 25cm to 20cm, which is exactly our robot\rq{}s width.  A field area of moderate weed densities was selected and the two DC motors moving two tracks of the robot were calibrated using PID parameters to have the same rotation speed, making the robot initially travel in a straight line without disturbances. The two spray tanks were filled with blue and red colored liquid in order for us to be able to evaluate the spraying performance by visual inspection. 

To examine the effect of inclement weather on the performance of the robot, we tested the robot on muddy soil after two days of intermittent rain in Fargo. It was evident that it was more difficult for the robot to move and much easier to get trapped on muddy soil because the mud stuck to the tracks. This indicates an opportunity for future work on the design of an all-weather robot.

\subsection{General Tests in Fields}
\label{test_in_field}

The tests were conducted with a continuous speed of $0.14$m$/$s when four-cell LiPo batteries were fully charged at $16.8$V, 10400mAh and $0.09$m$/$s when the voltage of two-cell battery was lower than $4.5$V. The batteries lasted more than 6 hours before requiring to be recharged. Under the guidance of contour algorithm, when the robot traveled in the aisles, both wide ones with a width of $25$cm, and narrow ones with an approximate width of $20$cm, it would not drive into croplines for the $3$m row length even when the field was full of uneven dents with a depth of $3-5$cm. 
The spraying algorithm worked remarkably in the actual fields. Even when the light conditions changed a lot during the day, the camera was able to recognize and spray weeds from 8 a.m. to 7:30 p.m. in Fargo. However, due to the delay we set after running a loop of spraying algorithm and communication time variation between the Arduino and Jetson Nano, the probability of spraying the same weed twice was 16.67\%. Effectiveness of the spraying system is verified in the submitted video. 


\subsection{Inter-row Navigation Algorithm Performance}
\label{sec:algorithm_result}

\textit{Prior-Otsu} method introduced in Section \ref{subsubsec:self-adjust} enabling the robot to adjust the HSV range by itself to relieve the effect of illumination on color-based algorithms. Experiments have shown that the method of updating HSV range through prior contour generally works well. Otsu's method is here as a backup when the hue value changes significantly. 
Fig. \ref{fig:otsu} clearly validates effectiveness of Otsu's method. \textit{Prior-Contour-Gradient}, our inter-row navigation algorithm, turns out to be experimentally effective while running the robot within 40 rows in the artificial fields on a university campus in Los Angeles, California under varying illumination conditions. The robot did not run into any (artificial) crops for 30 rows (1.5 meters per row) if the height of bumps in the path was less than 1.5cm, about half of the driving wheel of our robot, verifying the effectiveness of our inter-row navigation algorithm. However, if the row is longer than 3m or bumps' height are bigger than 1.5cm even with 1.5m row length, the integration of IMU is drifted, therefore the robot drives into crops because the current orientation of the robot is wrongly estimated by the IMU. The correction of integration error of the IMU is one of our future directions.

\begin{table}[t!]
\caption{Navigation algorithm performance comparison.}
\label{algorithm_accuracy}
\begin{center}
\begin{tabular}{|c||c||c||c||c|}
\hline
Algorithms & Accuracy & Precision & Recall & $F_{\textrm{score}}$\\
\hline
PHT  & 0.919 & 0.977 & 0.915 & 0.945\\
\hline
LSD &  0.806 & 0.927 & 0.809 & 0.864\\
\hline
FPE & 0.629 & 0.882 & 0.612 & 0.723\\
\hline
\textbf{Contour} &\textbf{0.935} & \textbf{0.978} & \textbf{0.938} & \textbf{0.957}\\
\hline
\end{tabular}
\end{center}
\end{table}

\begin{figure}[b!]
	\centering
	\includegraphics[width=\columnwidth]{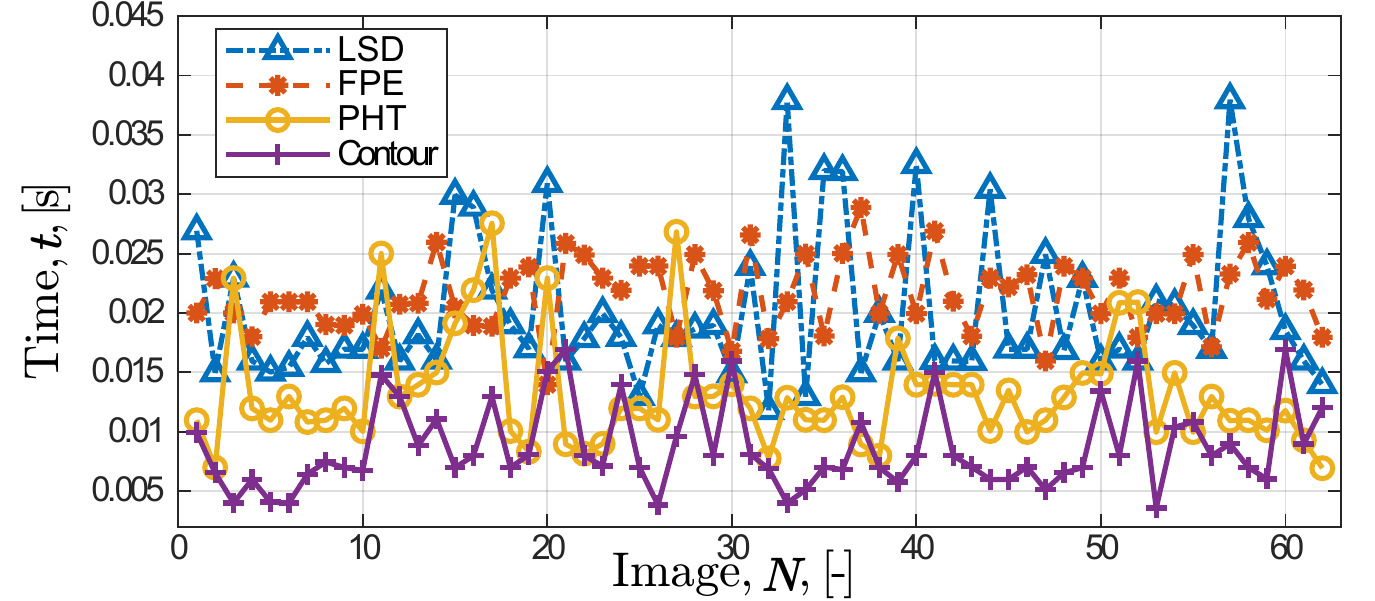}
	\caption{Run time comparison of LSD, FPE, PHT, and contour algorithm for each image. 
	}
	\label{fig:speed}
\end{figure}

The ground truth decision of the vanishing point is somewhat subjective. As shown in Fig. \ref{fig:preprocess}(b1-b4), the hollow circle denotes the ground truth of the vanishing point that is labeled by the researchers. The vanishing point (intersection of croplines) detected by algorithms is recognized as a true positive if it drops into the target circle centered at the ground truth with a radius of 5 pixels. The comparison of running speed for every frame among PHT, LSD, FPE and contour is shown in Fig.~\ref{fig:speed} and mean performance of the 62 frames is displayed in Table \ref{algorithm_accuracy}. Note that before each algorithm is run, our preprocessing techniques are used, without which the accuracy of existing line detection algorithms, i.e. PHT, LSD and FPE are going to be significantly reduced.
To quantitatively measure the efficacy, four indexes are chosen, accuracy, precision, recall and a combined $F_{\textrm{score}}$. 

 It turns out that contour algorithm, in the context of navigation in row crop field, is the fastest and most efficient for the robot to navigate inside narrow crop spacing. 

\subsection{Recharging System Robustness Test}
\label{subsec:rechargingExp}

\begin{figure*}[ht]
	\centering
	\includegraphics[width=0.65\textwidth]{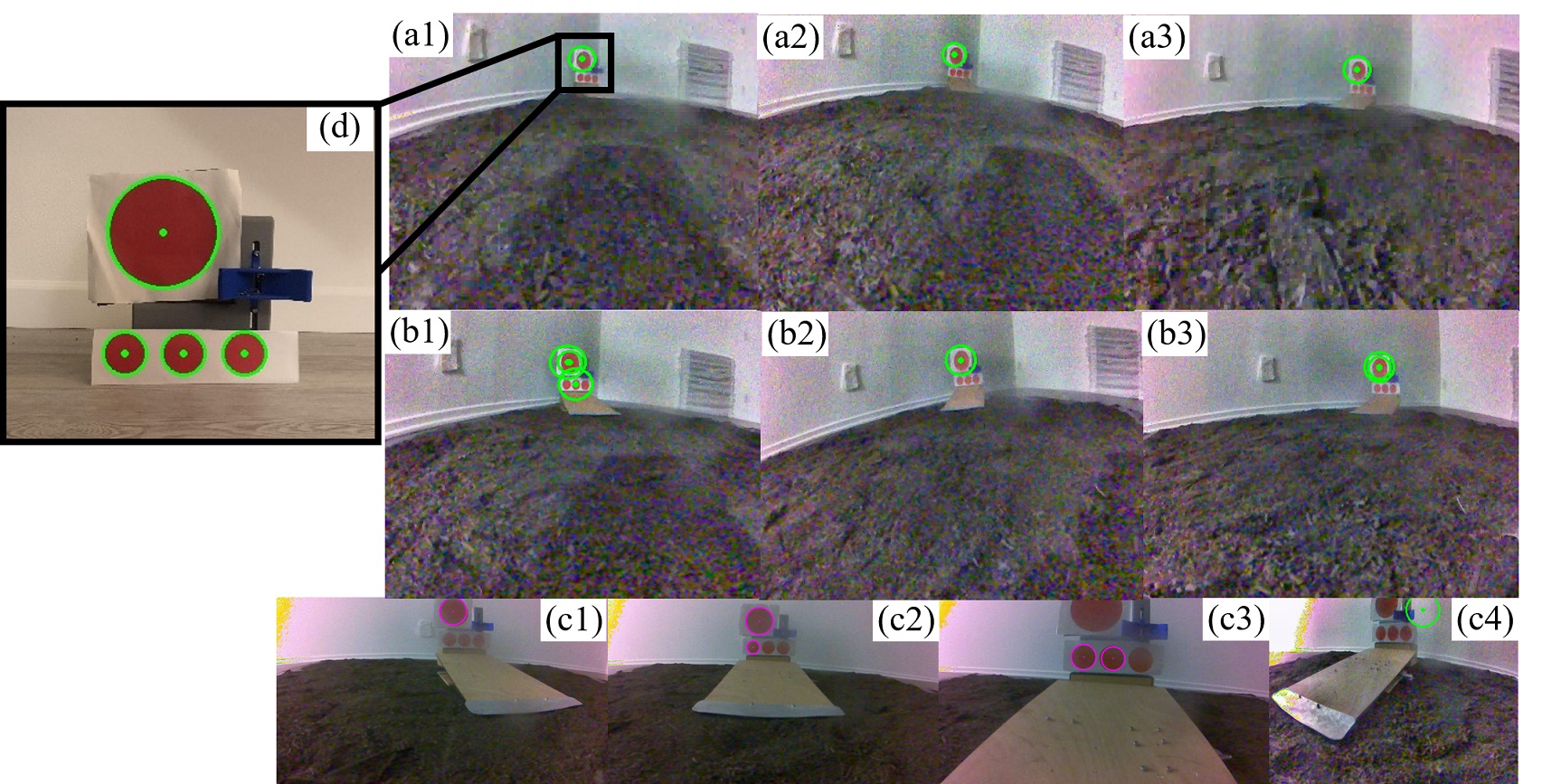}
	\caption{Images extracted from real-time video streaming on the robot camera demonstrate the detection of the circle pattern at different distances from the charging station from various orientations in the field. (a-c) The robot is roughly (a) 3.1, (b) 2.0, and (c) 0.9 meters from the charging station with a (1) 0.5 meter offset orientation on the left, (2) facing straight at the robot, and (3) 0.5 meter offset on the right. Note that the circles in (c1-c3) are detected by the embedded circle Hough algorithm in OpenCV while (c4) displays the circle detected by the mathematical definition of circles. 
	Inset: Pattern of circles on the charging station to help with alignment. (d) The closeup of red circle pattern and detection results with our \textit{Def-circle} algorithm (green dots are circle centers and lines are perimeters).
	}
	\label{fig:exp_circleDetection}
\end{figure*}

Fig. \ref{fig:ramp_in_fields} is extracted from the supplementary video recorded in the field where the recharging function was verified. Additionally, the \textit{Def-circle} algorithm  that is explained in Section \ref{subsubsec:rechargVision} is validated experimentally. Fig. \ref{fig:exp_circleDetection} gives a group of images exacted from the the real-time video streaming, displaying various camera views and corresponding circle detection results of the three-red-circle pattern. 
We performed 30 trials with the robot starting 3m away and heading towards the recharging station under the guidance of \textit{Def-circle} vision algorithm. It turns out that the robot recharged successfully 25 times. As a result, we conclude from experimental data that the robot can realize 85\% robust charging with our \textit{Def-circle} algorithm. The main causes of failure are the flare glow effect, which can be seen apparently from some images in Fig. \ref{fig:exp_circleDetection}, especially in (c4). Fig. \ref{fig:exp_circleDetection}(c4) is referred to as a `failed' case since the detected circle is not part of the circle pattern. Nonetheless, it will still lead the robot onto the ramp if the ramp is wide enough and if the robot is getting closer and closer to the station, it can be corrected from time to time.
To sum up, the recharging system is robust but still has room to improve. The inaccuracy due to the optical overexposure of the camera is a technical challenge for cost-effective onboard cameras. Besides, it might be possible that the parameters that need to be tuned by humans in our Def-cricle algorithm as mentioned above in Section \ref{subsubsec:rechargVision}, $numPts$, $\texttt{maxOfs}$, $minR$ and $maxR$, and $thres$ can be updated adaptively according to the distance from the charging station with more sensors added in the future. However, during our experiments, they are all constants ($numPts = 20$, $\texttt{maxOfs = 100}$, $minR = 10$ and $maxR = 100000$, and $thres: {H = \{ 0-30 \} \cup \{ 160- 180 \}, S = 0-255, V = 0-255}$) with an 85\% recharging accuracy as discussed. The accuracy will be improved after adding a sun protection visor onto the camera.

In summary, our preprocessing techniques together with contour, i.e. Pre-Contour algorithm, makes the crop row detection speedy and accurate. Cooperating with the gradient algorithm, it is finally called \textit{Pre-Contour-Gradient}, which is also capable of directing the robot\rq{}s turning. Overall, rapid and robust navigation has been achieved.

\section{Conclusions and Future Directions}
\label{sec:conclusion}
This paper put forth our design of a small and compact low-cost autonomous weed spraying robot with tracks. The total cost of the robot is less than $\$400$. Tested in flaxseed fields in North Dakota, the paramount flax-derived oil and food producer in the United States, it successfully realized computer vision based weed detection in narrow spacing fields, completed inter and intra-row navigation and automatic recharge. The robot can be used for autonomous weed control without human supervision. Given the modularity and the functionality, the agriculture robot can also be operated inside fields taking images of different kinds of weeds and building a massive weed library, based on which machine learning algorithms can be explored for detecting weeds within croplines. This can help provide valuable data on the effect of weed density on crop yield, and advance precision agriculture. Ultimately, this technology can support sustainable agriculture by reducing herbicide use by more than $90\%$~\cite{timmermann}.
Overall, the three contributions mentioned at the end of Section \ref{sec:introduction} are illustrated and verified. 

Although the experimental results are exciting, they also reveal pathways for improvement:

\begin{itemize}
	\item For hardware, the wooden ramp will be fabricated to be wider so that the robot is able to reach the charging point from all directions. In addition, the spraying system utilizes electronic misters that have low resistance to the wind power in real fields, so it will be updated. 
	
	\item Currently, the robot is not able to locate itself accurately in the field; this may be a concern is very large fields. Further research is needed on improved information fusion algorithms using the output from IMU, motor encoders, and cameras to implement simultaneous localization and mapping (SLAM) of the field. Also, one camera for multiple visual tasks may be inconvenient from a software development perspective; more cameras will be mounted onto the robot, thanks to their low cost. The video stream from one camera will be used for training an end-to-end model for weed-crop differentiation.
\end{itemize}



\section*{Acknowledgments}
We acknowledge support from the National Science Foundation (Award \# IIS - 1925360) and the Henry Samueli School of Engineering and Applied Science, University of California, Los Angeles.




\end{document}